\documentclass[12pt]{article}

\usepackage{textcomp}
\usepackage{booktabs}
\usepackage{amsmath}
\usepackage{multirow}
\usepackage{algorithm}
\usepackage{algorithmicx}
\usepackage{algpseudocode}
\usepackage{adjustbox}
\usepackage{placeins}
\usepackage{rotating}
\usepackage{natbib}
\usepackage{parskip}
\usepackage{url} 

\newcommand{\blind}{0}

\begin{document}

\def\spacingset#1{\renewcommand{\baselinestretch}%
{#1}\small\normalsize} \spacingset{1}


\if0\blind
{
  \title{\bf Pruning a neural network using Bayesian inference}
  \author{Sunil Mathew \\
    Department of Mathematical and Statistical Sciences, Marquette University\\
    and \\
    Daniel B.\ Rowe \\
    Department of Mathematical and Statistical Sciences, Marquette University}
  \maketitle
} \fi

\if1\blind
{
  \bigskip
  \bigskip
  \bigskip
  \begin{center}
    {\LARGE\bf Pruning a neural network using Bayesian inference}
\end{center}
  \medskip
} \fi

\bigskip
\begin{abstract}
  Neural network pruning is a highly effective technique aimed at reducing the computational and memory demands of large neural networks. In this research paper, we present a novel approach to pruning neural networks utilizing Bayesian inference, which can seamlessly integrate into the training procedure. Our proposed method leverages the posterior probabilities of the neural network prior to and following pruning, enabling the calculation of Bayes factors. The calculated Bayes factors guide the iterative pruning. Through comprehensive evaluations conducted on multiple benchmarks, we demonstrate that our method achieves desired levels of sparsity while maintaining competitive accuracy.
\end{abstract}

\noindent%
\vfill

\newpage
\spacingset{1.5} 
\section{Introduction}
\label{sec:intro}

In artificial neural networks (ANN) and machine learning (ML), parameters represent what the network has learned from the data. The number of parameters in a neural network can determine its capacity to learn.  With advancements in hardware capabilities, we can now define larger models with millions of parameters. The ImageNet Large Scale Visual Recognition Challenge (ILSVRC) and its winners over the years demonstrate how the error rate has decreased with an increase in the number of parameters and connections in neural networks. For instance, in 2012, AlexNet \citep{AlexNet}, one of the convolutional neural networks (CNNs), had over 60M parameters. The large language model, Generative Pre-trained Transformer 3 (GPT-3) \citep{brown2020language}, comprises 175 billion parameters. Even though deep neural networks with large number of parameters capture intricate underlying patterns, the large number of connections can introduce computational challenges, overfitting, and lack of generalizability. To address these issues, various methods have been developed.

Neural network pruning is a widely used method for reducing the size of deep learning models, thereby decreasing computational complexity and memory footprint \citep{lecun1989optimal, han2015learning, liu2018rethinking}. Pruning is crucial for deploying large models on resource-constrained devices such as personal computers, mobile phones and tablets. Pruning can also be used to reduce the carbon footprint of deep learning models by reducing the computational requirements \citep{strubell2019energy}. Pruning can also be used to improve the interpretability of deep learning models by removing redundant neurons or connections \citep{han2015learning}. 

Pruning methods can be classified into mainly three categories, weight pruning, neuron pruning, and filter pruning \citep{han2015deep, srivastava2014dropout, li2017pruning, he2018channel}. Weight pruning involves removing individual weights from the network based on their magnitude or other criteria, neuron pruning and filter pruning involve removing entire neurons or filters that are not important. Even though pruning methods can effectively reduce network size and improve performance, they often lack a principled approach for selecting the most important weights or neurons \citep{blalock2020state}.

In Bayesian neural networks, the weights of the network are treated as random variables with a prior distribution, which can be updated to get a posterior distribution using Bayes' rule. It allows us to quantify the uncertainty associated with each weight and select the most important weights based on their relevance to the task the network is being trained for. The posterior distribution reflects our updated belief about the weights based on the observed data and can be used to calculate the probability of each weight being important for the task at hand. Variational inference, which involves minimizing the Kullback-Leibler (KL) divergence between the true posterior and an approximate posterior, is a common approach for approximating the posterior distribution for neural network pruning \citep{dusenberry2019bayesian, blundell2015weight}. Other approaches include Monte Carlo methods and Markov chain Monte Carlo (MCMC) sampling \citep{molchanov2019variational}. However, these methods are computationally expensive and can prove to be difficult to be scaled to large networks.

In this work, we propose a Bayesian pruning algorithm based on Bayesian hypothesis testing. It provides a principled approach for pruning a neural network to a desired size without sacrificing accuracy. We compare two neural network models at every training iteration, the original unpruned network, and the pruned network. This comparison helps us to determine which model fits the data better. The ratio of the posterior probabilities of the pruned network to the posterior probabilities of the unpruned network (Bayes factor) can be then used to determine whether to prune the network further or skip pruning at the next iteration. This approach enables us to implement this method in regular neural networks without the need for additional parameterization as in the case of Bayesian neural networks.

\subsection{Pruning Neural Networks using Bayesian Inference}

\FloatBarrier

\begin{figure}[!htbp]
  \centering
  \includegraphics[width=0.8\textwidth]{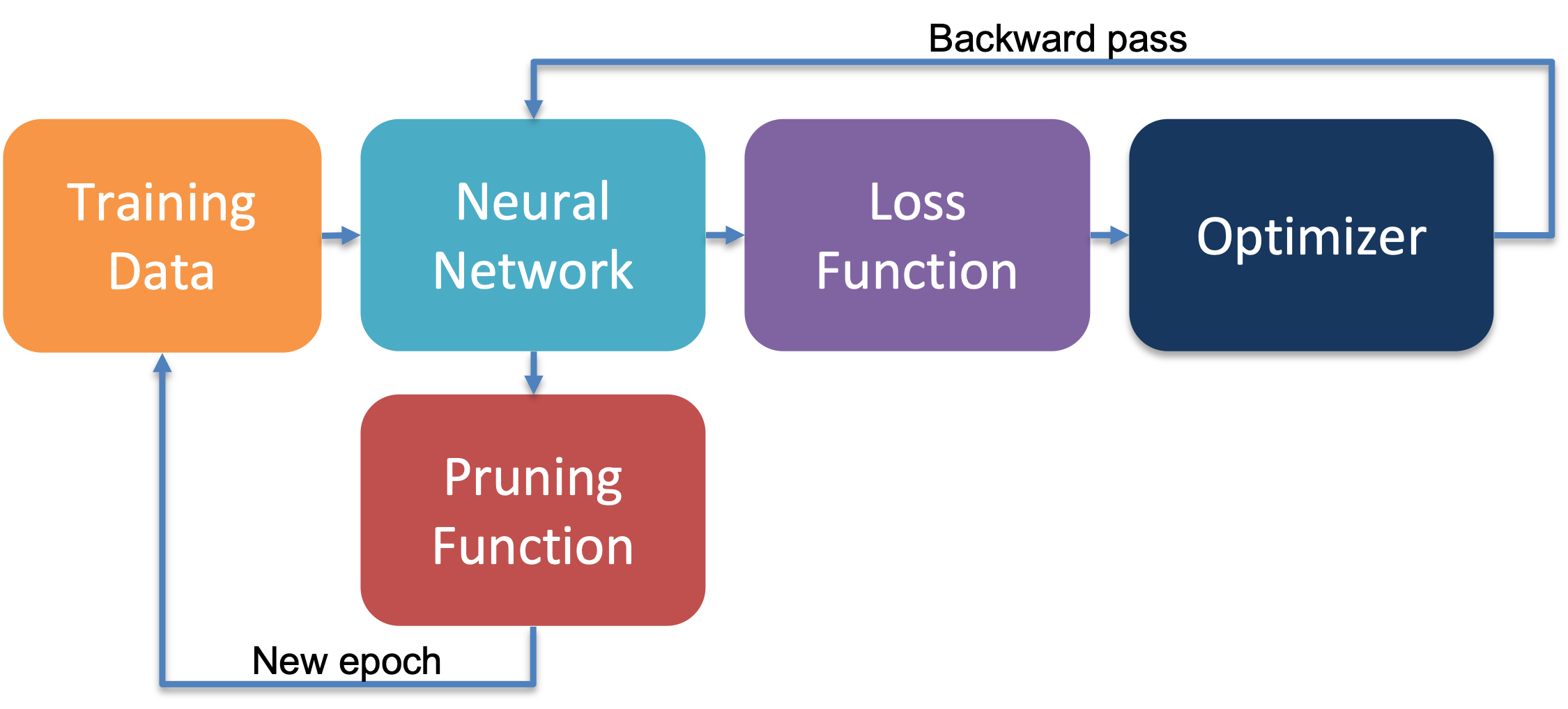}
  \caption{Pruning system block diagram.}
  \label{fig:pruning_system_block}
\end{figure}

The pruning system, seen in Figure \ref{fig:pruning_system_block}, incorporates pruning into the training process. The training data is divided into batches and processed by the neural network through a forward pass, consisting of matrix multiplications and non-linear activations. The network's output is then compared with the ground truth labels to compute the loss. The weights of the network is adjusted through a backward pass using an optimizer such as Stochastic Gradient Descent (SGD) or Adam \citep{kingma2015adam}. After each epoch, the weights are pruned using the pruning algorithm, and the pruned weights are used in the subsequent epochs. The pruning algorithm is based on Bayesian hypothesis testing, which is a statistical framework that can be used to compare two models, two network configurations in this case, to determine which one fits the data better. 

\FloatBarrier

To test the hypothesis that the pruned network fits the data better than the unpruned network, we define the null hypothesis as the unpruned network fitting the data better ($\theta = \psi$) and the alternative hypothesis as the pruned network fitting the data better ($\theta = \phi$). The Bayes factor, which is the ratio of the posterior probability of the alternative hypothesis to the posterior probability of the null hypothesis, is computed as follows:

$$\text{Bayes factor} = \frac{P(\theta = \phi | D)}{P(\theta = \psi | D)}$$

Here, $D$ represents the training data.

The posterior probability of the null hypothesis ($P(\theta = \psi | D)$) is computed as:

$$P(\theta = \psi | D) = \frac{P(D | \theta = \psi) P(\theta = \psi)}{P(D)}$$

Similarly, the posterior probability of the alternative hypothesis ($P(\theta = \phi | D)$) is computed as:

$$P(\theta = \phi | D) = \frac{P(D | \theta = \phi) P(\theta = \phi)}{P(D)}$$

The Bayes factor is then calculated as the ratio of the posterior probabilities:

$$\text{Bayes factor} = \frac{P(D | \theta = \phi) P(\theta = \phi)}{P(D | \theta = \psi) P(\theta = \psi)}$$

A Bayes factor greater than 1 indicates that the pruned network fits the data better, while a value less than 1 indicates that the unpruned network fits the data better.

For a classification problem, the likelihood of the data is given by the categorical cross-entropy loss function:

\begin{equation*}
  \log p(y_{pred} | y_{true}) = \log \mathcal{C}(\mathrm{softmax}(y_{pred})){y_{true}}
\end{equation*}

Here, $y_{\text{pred}}$ represents the neural network's predictions for the classes, and $y_{\text{true}}$ is the ground truth. A Gaussian prior with mean $\mu$ and variance $\sigma^2$ is used for weights: 

\begin{equation*}
p(w) = \mathcal{N}(\mu, \sigma^2)
\end{equation*}

The log prior and log likelihood for the weight parameters are used to compute the log posterior distribution of the weights:

\begin{equation*}
    \log p(w | D) = \log p(D | w) + \log p(w) \\
\end{equation*}

The log posterior is calculated before and after weight pruning to compute the Bayes factor. If the Bayes factor exceeds a predefined threshold, a certain percentage ($r$) of the weights are pruned as, 

\begin{equation}
w_{\text{new}} = w_{\text{old}} \odot m
\label{eq:pruning}
\end{equation}

where $\odot$ represents element-wise multiplication, $w_{\text{old}}$ is the old weight matrix, and ${m}$ is the binary mask indicating which weights should be pruned (i.e., have a value of 0) and which weights should be kept (i.e., have a value of 1). The resulting matrix $w_{\text{new}}$ has the same dimensions as $w_{\text{old}}$, but with some of its weights pruned.
Algorithm \ref{alg:bayes_pruning} outlines the Bayesian pruning process.

\begin{algorithm}[h!]
  \caption{Bayesian Pruning Algorithm}
  \label{alg:bayes_pruning}
  \begin{algorithmic}[3]
  \State Input: Trained neural network $f(\cdot,w)$, pruning rate $r$, dataset $\mathcal{D}={(\mathbf{x}i, y_i)}_{i=1}^n$, $\beta$ Bayes factor threshold
  \ Output: Pruned neural network $f_{r}(\cdot,w)$
  \State Compute the posterior probability of the weights before pruning
  \If{$BF_{01}>\beta$}
    \State Prune $r$ percentage of weights of $f(\cdot,w)$
    \State Return $f_{r}(\cdot,w)$
  \EndIf
  \State Compute the posterior probability of the weights after pruning
  \State Compute the Bayes factor using the posterior probabilities before and after pruning
  \end{algorithmic}
\end{algorithm}

\FloatBarrier

In the following sections, we introduce two pruning algorithms that utilize this framework: random pruning, which randomly selects weights for pruning, and magnitude pruning, which prunes weights based on their magnitude.

\subsubsection*{Bayesian Random pruning}

Random pruning is a simple pruning algorithm that randomly selects weights to prune. Here we set the pruning rate to be the desired level of sparsity that we are looking to achieve. After an epoch, we count the number of non-zero parameters in the network and randomly zero out just enough parameters to achieve the desired level of sparsity. The algorithm is summarized in Algorithm \ref{alg:random_pruning}.

\begin{algorithm}[h!]
  \caption{Bayesian Random Pruning}
  \label{alg:random_pruning}
  \begin{algorithmic}[1]
  \State $f(\cdot,w)$: Neural network model with parameters $w$
  \State $r$: Desired sparsity level, $\beta$ Bayes factor threshold
  \State Calculate log posterior probability $p(w | \mathcal{D})$
  \If {$BF_{01} > \beta$}
  \ForAll{weights $w_i \in w$}
      \State $n \leftarrow \text{size}(w_i)$
      \State number of weights to prune, $k \leftarrow (n \times r)$
      \State $I \leftarrow \text{indices \ of \ non zero weights}$
      \State $n_z \leftarrow \text{number of zero weights}$
      \State $k' \leftarrow k - n_z$
      \State $J \leftarrow \text{random\_sample}(I, k')$
      \State set elements in $w_i$ at indices $J$ to zero
  \EndFor
  \EndIf
  \State Calculate log posterior probability $p(w | \mathcal{D})$ after pruning
  \State Calculate Bayes factor $BF_{01}$
  \end{algorithmic}
\end{algorithm}

\FloatBarrier

\subsubsection*{Bayesian Magnitude pruning}

Magnitude pruning is a pruning algorithm that selects weights to prune based on their magnitude. This can be seen as pruning weights that are less important. Here we set the pruning rate to be the desired level of sparsity that we are looking to achieve. The lowest weights corresponding to the desired level of sparsity is pruned to get the pruned network. The algorithm is summarized in Algorithm \ref{alg:magnitude_pruning}.

\begin{algorithm}[h!]
  \caption{Bayesian Magnitude Pruning}
  \label{alg:magnitude_pruning}
  \begin{algorithmic}[1]
  \State $f(\cdot,w)$: Neural network model with parameters $w$
  \State $r$: Desired sparsity level, $\beta$ Bayes factor threshold
  \State Calculate log posterior probability $p(w | \mathcal{D})$
  \If {$BF_{01} > \beta$}
  \ForAll{weights $w_i \in w$}
      \State $n \leftarrow \text{size}(w_i)$
      \State number of weights to prune, $k \leftarrow (n \times r)$
      \State $w_i \leftarrow sort(w_i)$
      \State set $k$ elements in $w_i$ to zero
  \EndFor
  \EndIf
  \State Calculate log posterior probability $p(w | \mathcal{D})$ after pruning
  \State Calculate Bayes factor $BF_{01}$
  \end{algorithmic}
\end{algorithm}

\FloatBarrier

\subsection*{Experimental Setup}

To evaluate the performance of Bayesian Random Pruning and Bayesian Magnitude Pruning, we conduct experiments on three datasets and two neural network architectures for five different levels of desired sparsity. The datasets used arne MNIST \citep{lecun1998gradient}, MNIST Fashion \citep{xiao2017fashion} and CIFAR-10 \citep{krizhevsky2009learning}. The neural network architectures are a Fully Connected Network (FCN) and a Convolutional Neural Network (CNN). The five different levels of sparsity are 25\%, 50\%, 75\%, 90\% and 99\%. We use a learning rate of 0.001 and a batch size of 64 for all experiments. Data preprocessing only consist of normalizing the dataset and does not include any data augmentation like Random cropping or flipping of images to have less confounding variables in the studies we conduct to observe the effects of our pruning algorithm. We train the network for 25 epochs on the training set and evaluate its performance on the test set. We evaluate the performance of each method in terms of the accuracy of predictions it makes for the target classes using the test set. Each experiment is repeated 5 times and the mean and standard deviation of the accuracy is reported. 

The following sections describe the neural network architectures used in our experiments.

\FloatBarrier

\subsection*{Neural Network Architectures}

The two neural network architectures used in our experiments are the Fully Connected Network (FCN) and the Convolutional Neural Network (CNN). The same architectures are used for all three datasets. The FCN consists of two hidden layers. The output of the last fully connected layer is fed into a softmax layer to get the class probabilities. The CNN consists of two convolutional layers with 32 and 64 filters respectively followed by two fully connected layers. Each convolutional layer is followed by a max pooling layer with a kernel size of 2 and stride of 2. The output of the second max pooling layer is flattened and fed to the fully connected layers. The output of the fully connected layer is fed into a softmax layer to get the class probabilities. 

\begin{figure}[h!]
  \centering
  \begin{adjustbox}{width=0.6\textwidth, height=0.6\textwidth}
  \includegraphics{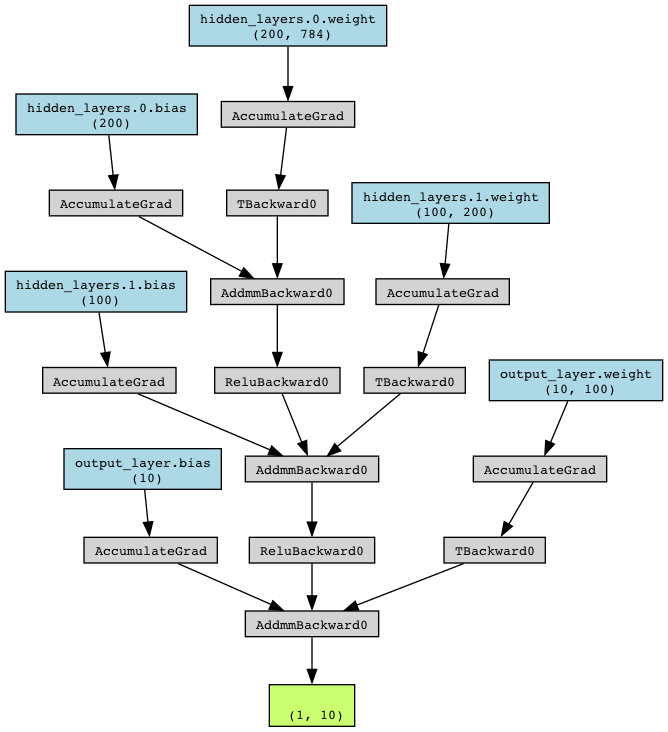}
  \end{adjustbox}
  \caption{Fully connected neural network architecture}
  \label{fig:fcn}
\end{figure}

The network architecture of the fully connected network (FCN) is seen in Figure \ref{fig:fcn}. 
\FloatBarrier



\begin{figure}[h!]
\centering
\begin{adjustbox}{width=0.6\textwidth, height=0.6\textwidth}
\includegraphics{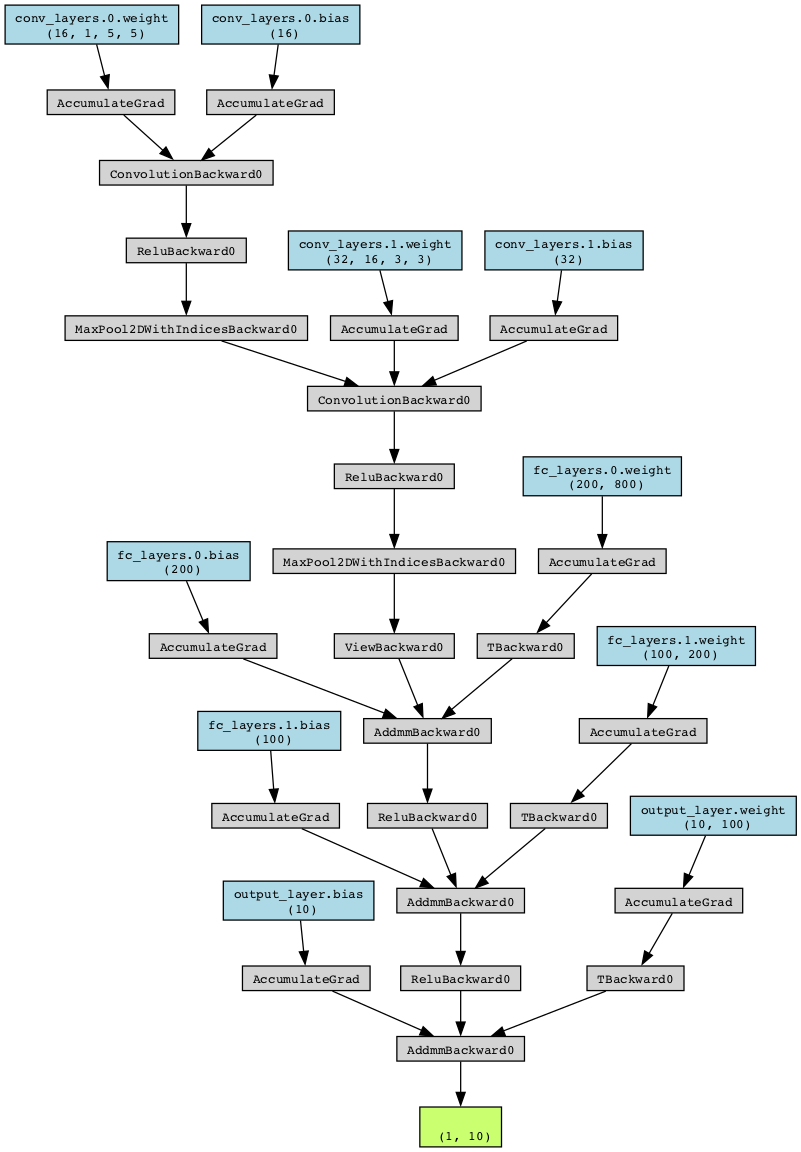}
\end{adjustbox}
\caption{Convolutional neural network architecture}
\label{fig:cnn}
\end{figure}

The network architecture of the convolutional neural network (CNN) is seen in Figure \ref{fig:cnn}. 

\FloatBarrier

\FloatBarrier

\subsection{Results}

The following sections present the results of the experiments. The results are presented in the following order: (1) MNIST dataset, (2) MNIIST-Fashion dataset, and (3) CIFAR-10 dataset. The results are presented in the form of learning curves and a table with accuracy for different levels of sparsity for the FCN and CNN model. The accuracy is the percentage of correctly classified images in the test set. The sparsity is the percentage of weights that are pruned in the network.  The results are compared to baseline, which is the model trained without pruning, and the non-Bayesian version of the pruning method.

\subsection*{MNIST}

\begin{figure}[h!]
\centering
\includegraphics[width=\textwidth]{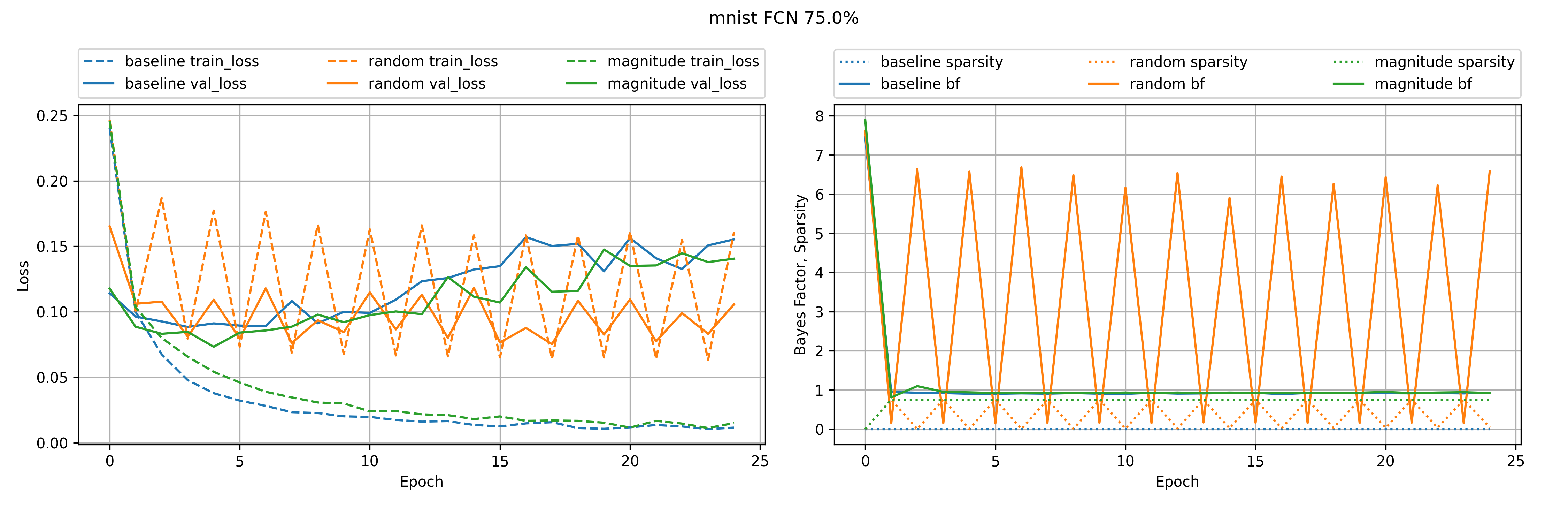}
\caption{MNIST (FCN 75\%) learning curves for the Bayesian pruning method.}
\label{fig:mnist-learning-curve_fcn}
\end{figure}

Figure \ref{fig:mnist-learning-curve_fcn} shows the learning curves for random pruning, magnitude pruning under a Bayesian framework compared to baseline in a fully connected network (FCN) trained on the MNIST dataset. Here the desired level of sparsity is 75\%. The figure has two subplots. One shows the training and validation loss as a function of the number of epochs, the other plot (right) shows the Bayes factor, sparsity as a function of the number of epochs.

The training loss is the average loss over the training set, and the validation loss is the average loss over the validation set. The figure shows that the training loss decreases as the number of epochs increases, and the validation loss starts to decrease in about 5 epochs. The  training loss decreases faster than the validation loss, which indicates that the model is overfitting the training data. As pruning begins, it affects the training and validation loss of both random and magnitude pruning as seen the curves. There are large oscillations in loss values for random pruning as seen in the figure. The Bayes factor begins to reduce as the number of epochs increases and the sparsity of the network becomes stabilized for magnitude pruning, but it remains fluctuating for random pruning and shows an increasing trend for the Bayes factor suggesting that Bayesian random pruning fits the data better than other methods.

\begin{figure}[h!]
  \centering
  \includegraphics[width=\textwidth]{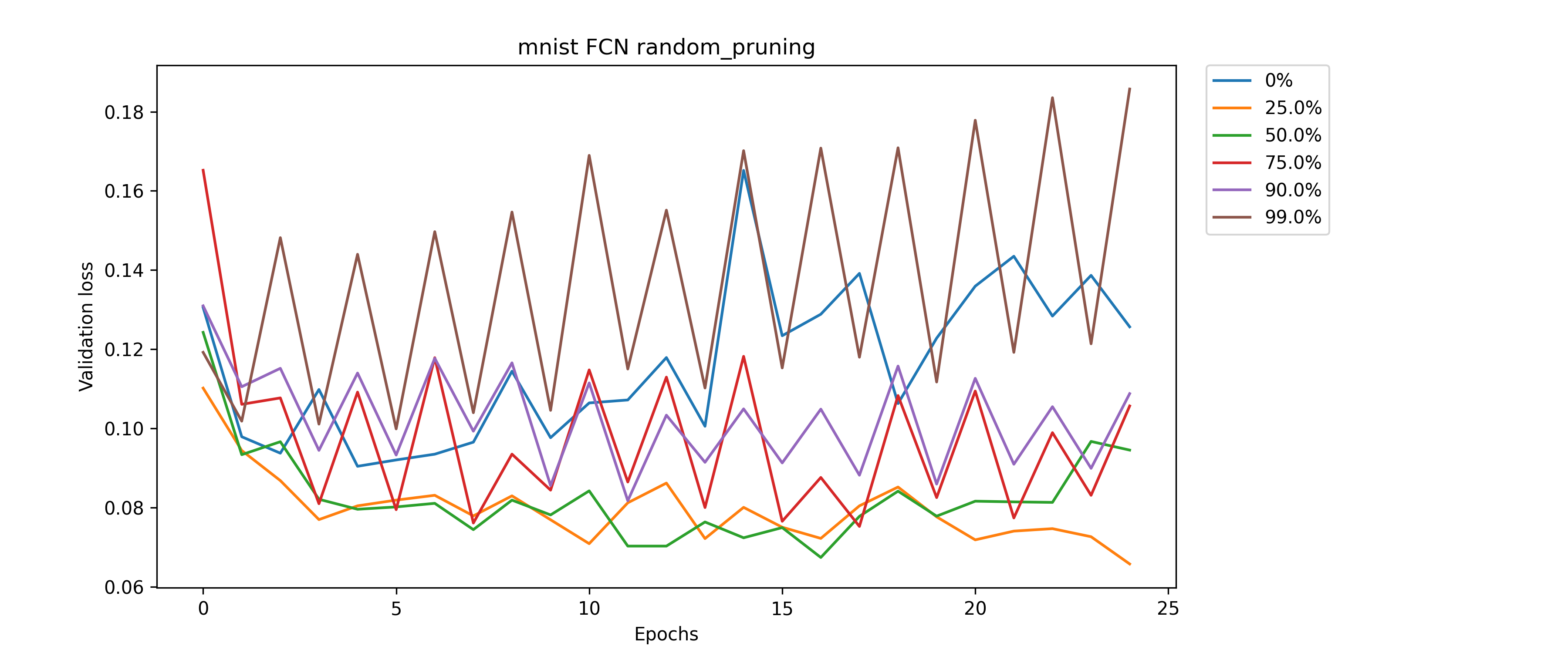}
  \caption{Validation loss of random pruning for different sparsity levels.}
  \label{fig:mnist_FCN_25_random_pruning_loss_vs_sparsity}
\end{figure}
  
Figure \ref{fig:mnist_FCN_25_random_pruning_loss_vs_sparsity} shows the validation accuracy of random pruning for different sparsity levels. For 25\% sparsity the validation accuracy seems to be the highest. Then as the sparsity level increases the validation accuracy begins to decrease. Until 90\% sparsity the validation accuracy remains to have a downward trend and combats overfitting compared to the baseline. The network only starts to become worse at 99\% sparsity.

\FloatBarrier
  
\begin{figure}[h!]
  \centering
  \includegraphics[width=\textwidth]{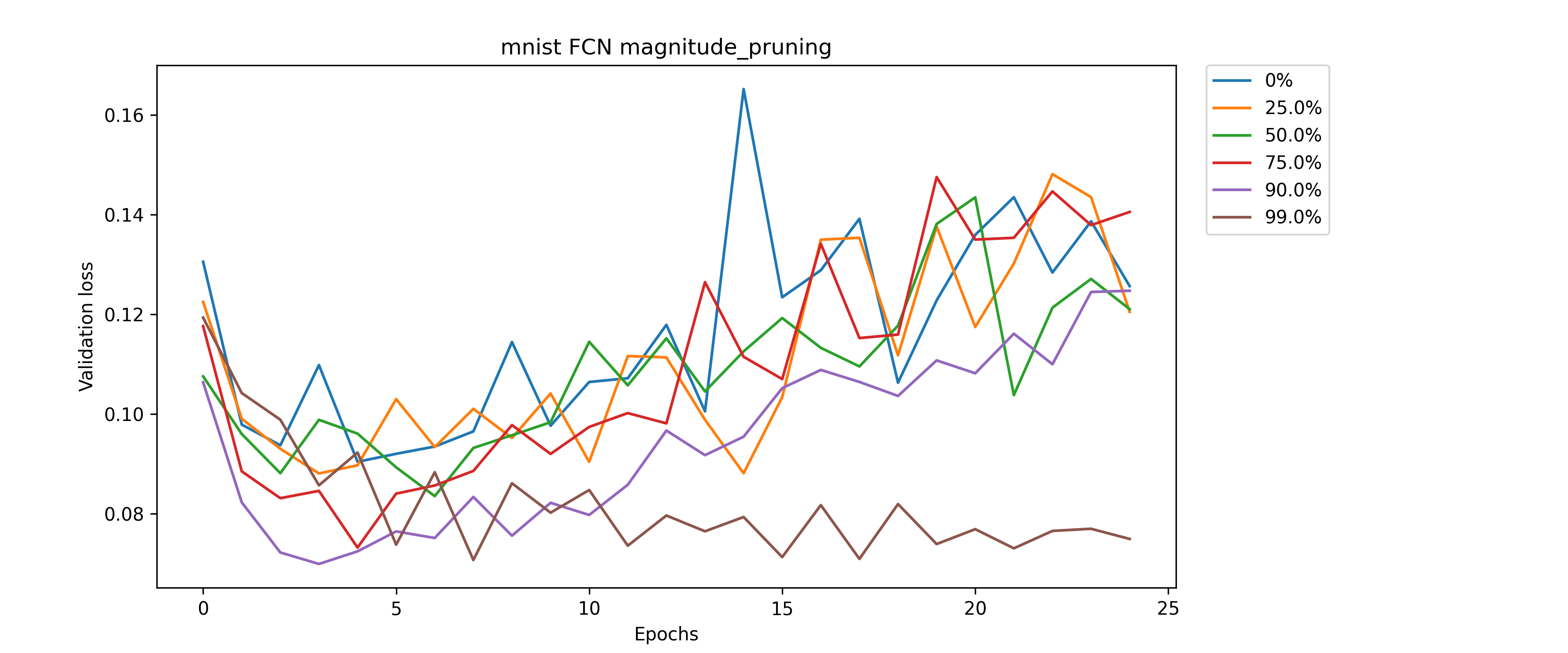}
  \caption{Validation loss of magnitude pruning for different sparsity levels.}
  \label{fig:mnist_FCN_25_magnitude_pruning_loss_vs_sparsity}
\end{figure}
  
Figure \ref{fig:mnist_FCN_25_magnitude_pruning_loss_vs_sparsity} shows the validation accuracy of magnitude pruning for different sparsity levels. For 25\% sparsity the validation accuracy remains similar to the baseline. Then as the sparsity level increases the validation accuracy starts to improve, but the network still overfits the data until 99\% of the parameters are pruned.

\begin{figure}[h!]
\centering
\includegraphics[width=\textwidth]{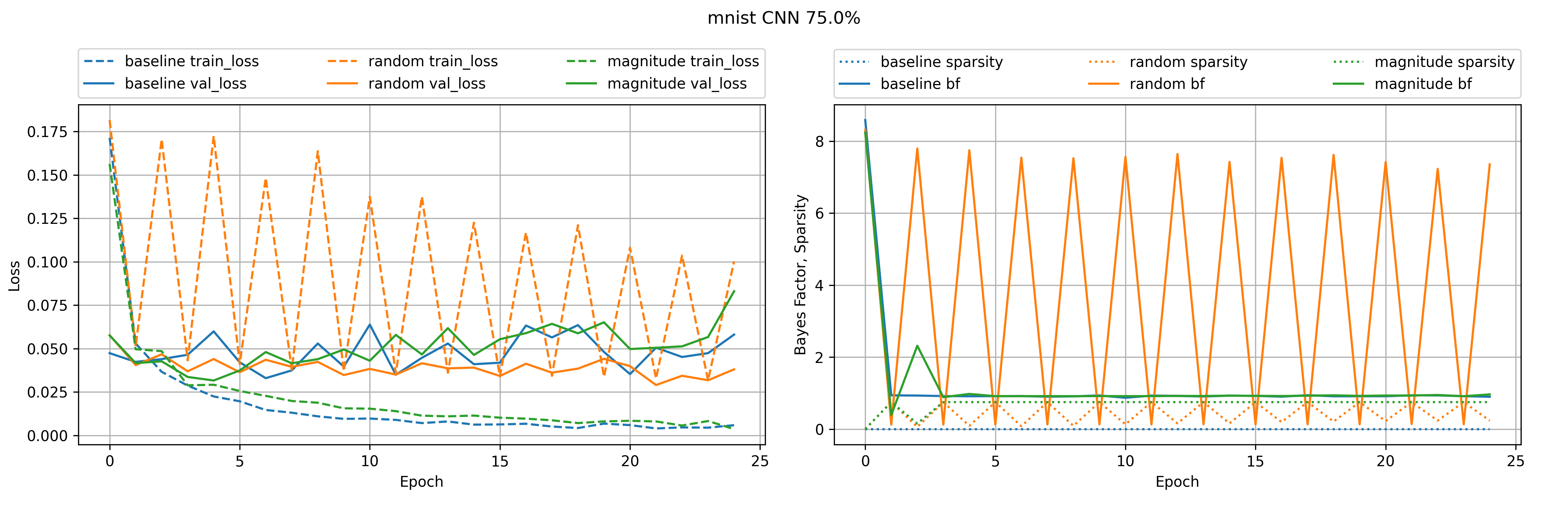}
\caption{MNIST (CNN 75\%) learning curves for the Bayesian pruning method.}
\label{fig:mnist-learning-curve_cnn}
\end{figure}

Figure \ref{fig:mnist-learning-curve_cnn} shows the learning curves for random pruning, magnitude pruning under a Bayesian framework compared to baseline in a convolutional neural network (CNN) trained on the MNIST dataset. The number of parameters in the CNN are comparatively larger than that of the FCN. This causes the effects of overfitting to be seen a little later in the training period and less overfitting compared to the FCN at 75\% sparsity. Bayes factor for random pruning is higher than that of magnitude pruning, which suggests that Bayesian random pruning fits the data better. 

\begin{figure}[h!]
\centering
\includegraphics[width=\textwidth, height=0.3\textwidth]{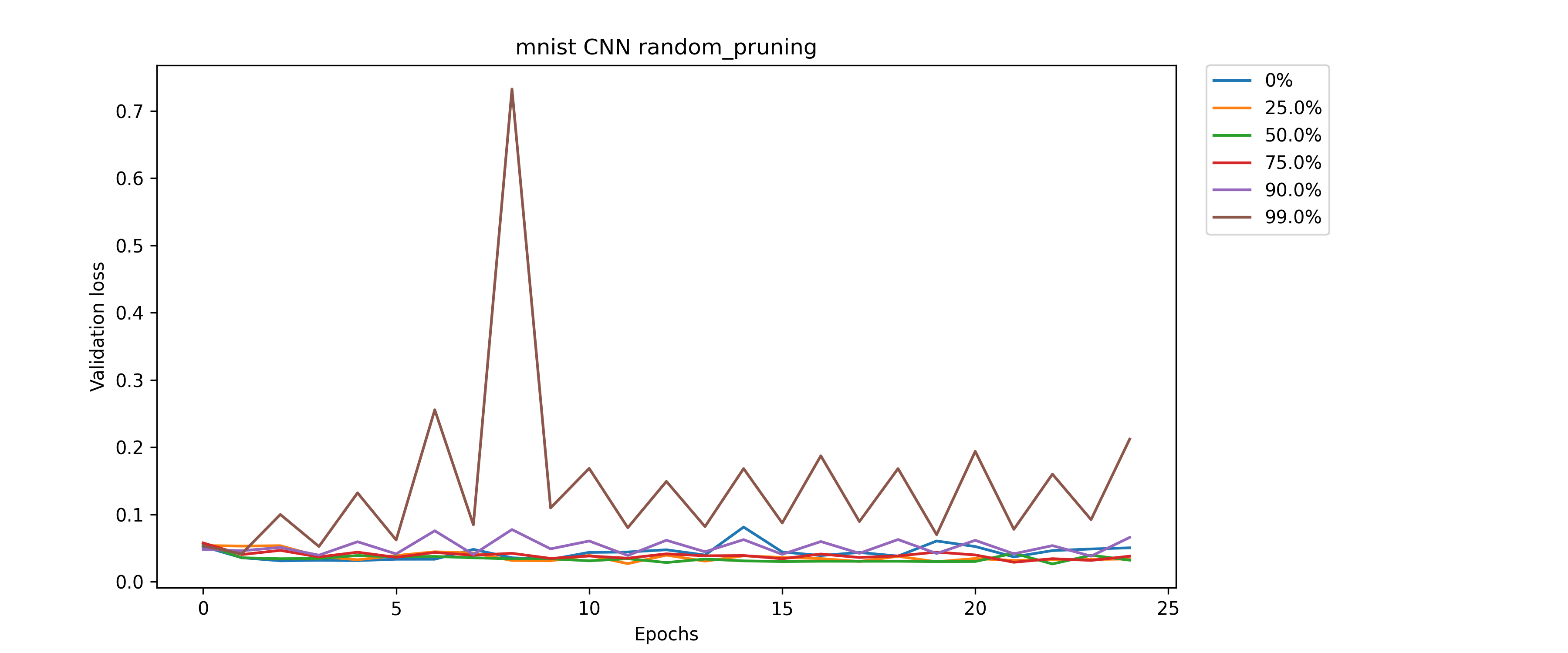}
\caption{Validation loss of random pruning for different sparsity levels.}
\label{fig:mnist_CNN_25_random_pruning_loss_vs_sparsity}
\end{figure}

\FloatBarrier

Figure \ref{fig:mnist_CNN_25_random_pruning_loss_vs_sparsity} shows the validation accuracy of random pruning for different sparsity levels. As the number of parameters of the CNN is larger than that of the FCN, the validation accuracy remains similar to the baseline until 90\% sparsity. Then as the sparsity level increases the validation accuracy begins to decrease.

\begin{figure}[h!]
\centering
\includegraphics[width=\textwidth]{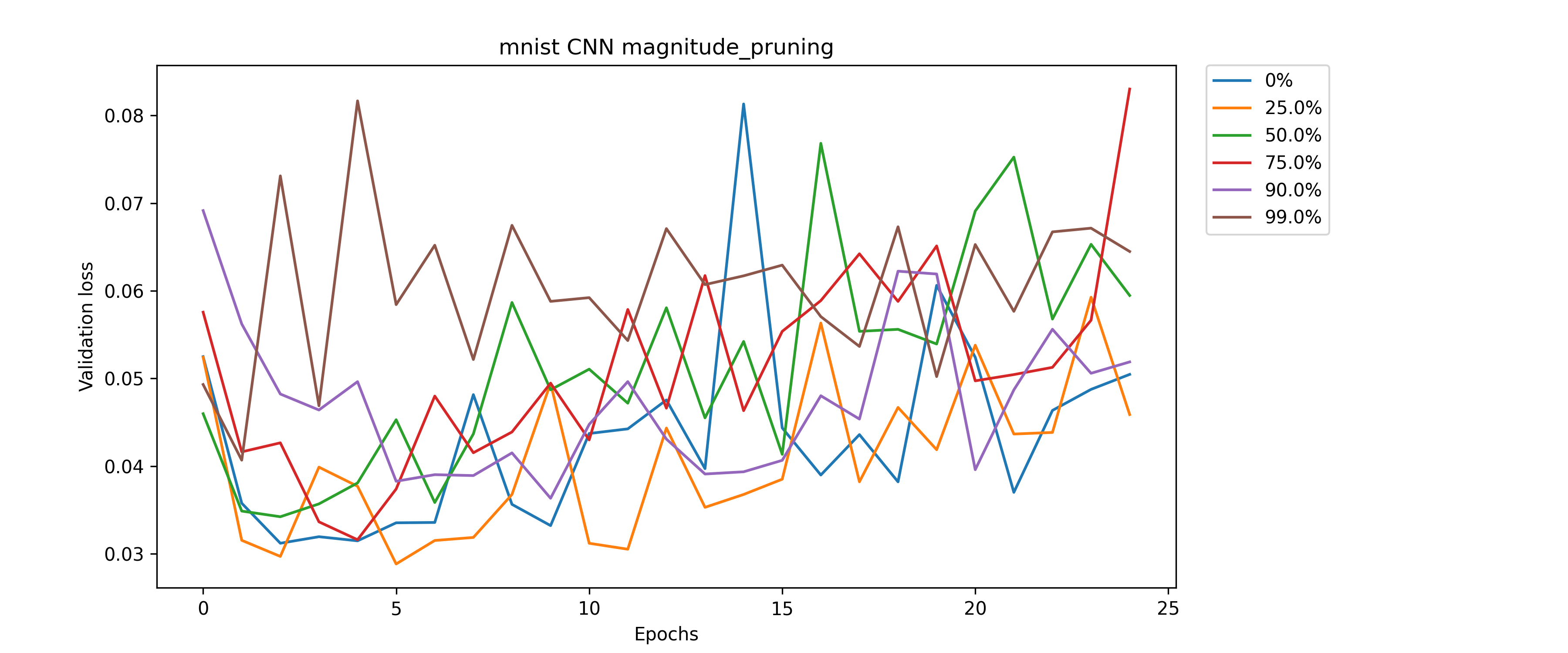}
\caption{Validation loss of magnitude pruning for different sparsity levels.}
\label{fig:mnist_CNN_25_magnitude_pruning_loss_vs_sparsity}
\end{figure}

\FloatBarrier

Figure \ref{fig:mnist_CNN_25_magnitude_pruning_loss_vs_sparsity} shows the validation accuracy of magnitude pruning for different sparsity levels. Even pruning 99\% of the parameters does not affect the validation accuracy of the CNN. This is because the CNN has an enormous number of parameters and the network overfits the data even after pruning 99\% of the parameters.

\subsection*{MNIST Fashion}

\begin{figure}[h!]
\centering
\includegraphics[width=\textwidth]{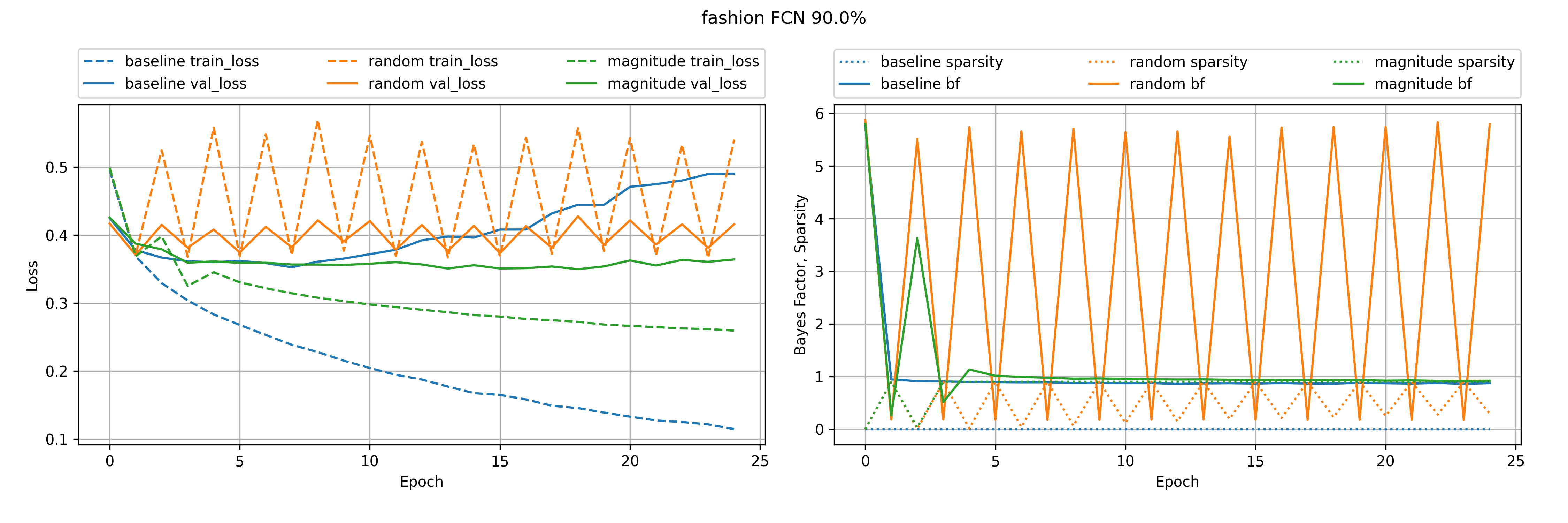}
\caption{MNIST-Fashion (FCN 90\%) learning curves for the Bayesian pruning method.}
\label{fig:fashion_FCN_25_0.90_loss}
\end{figure}

Figure \ref{fig:fashion_FCN_25_0.90_loss} shows the learning curves for random pruning, magnitude pruning under a Bayesian framework compared to baseline in a fully connected network (FCN) trained on the MNIST Fashion dataset. Here the desired level of sparsity is 90\%. The figure has two subplots. One shows the training and validation loss as a function of the number of epochs, the other plot (right) shows the Bayes factor, sparsity as a function of the number of epochs. 

\FloatBarrier

The training loss is the average loss over the training set, and the validation loss is the average loss over the validation set. The figure shows that the training loss decreases as the number of epochs increases, and the validation loss starts to decrease in about 5 epochs. The  training loss decreases faster than the validation loss, which indicates that the model is overfitting the training data. As pruning begins, it affects the training and validation loss of both random and magnitude pruning as seen the curves. There are large oscillations in loss values for random pruning. The Bayes factor begins to reduce as the number of epochs increases and the sparsity of the network becomes stabilized for magnitude pruning, but it remains fluctuating for random pruning. Bayesian random pruning model fits the data better than magnitude pruning model.

\begin{figure}[h!]
\centering
\includegraphics[width=\textwidth]{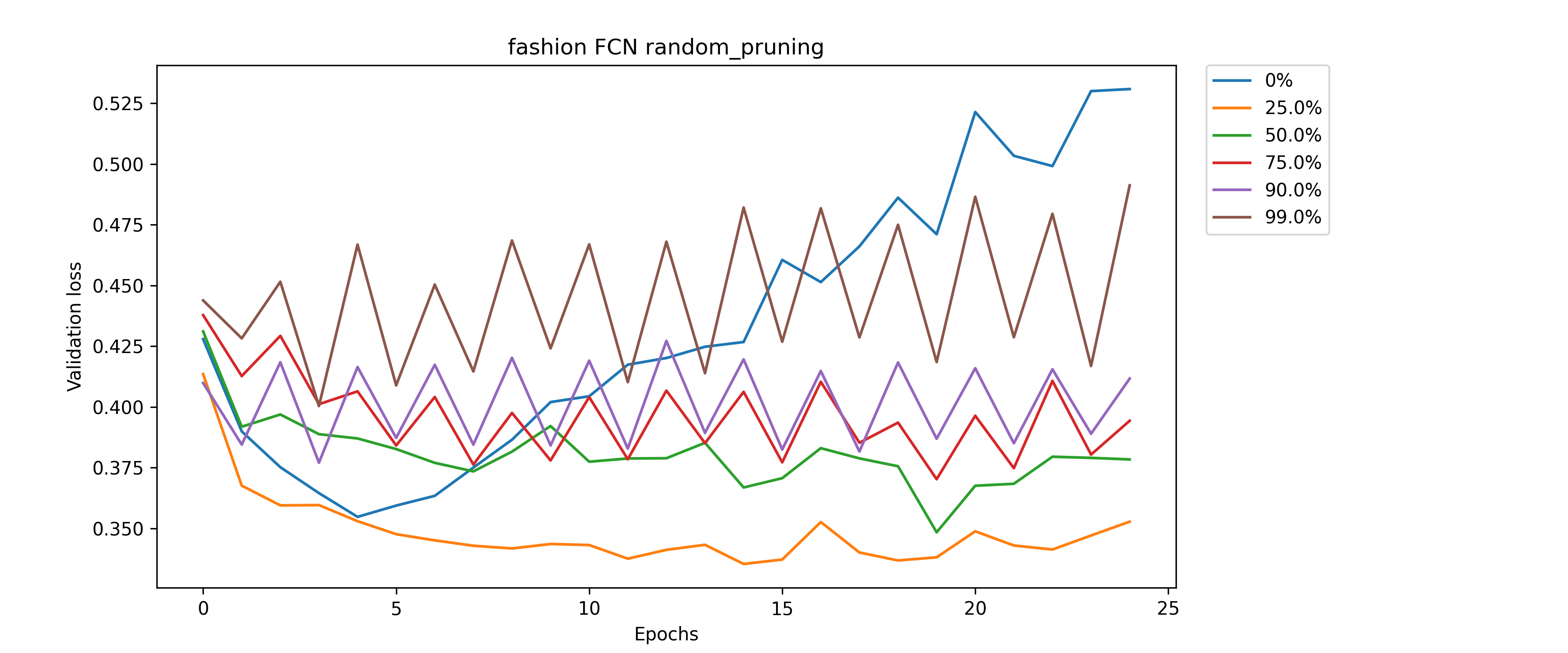}
\caption{Validation loss of random pruning for different sparsity levels.}
\label{fig:fashion_FCN_25_random_pruning_loss_vs_sparsity}
\end{figure}

\FloatBarrier

Figure \ref{fig:fashion_FCN_25_random_pruning_loss_vs_sparsity} shows the validation accuracy of random pruning for different sparsity levels. Similar to the MNIST dataset, the validation loss is the lowest for 25\% sparsity. Then as the sparsity level increases the validation accuracy begins to decrease.

\begin{figure}[h!]
\centering
\includegraphics[width=\textwidth]{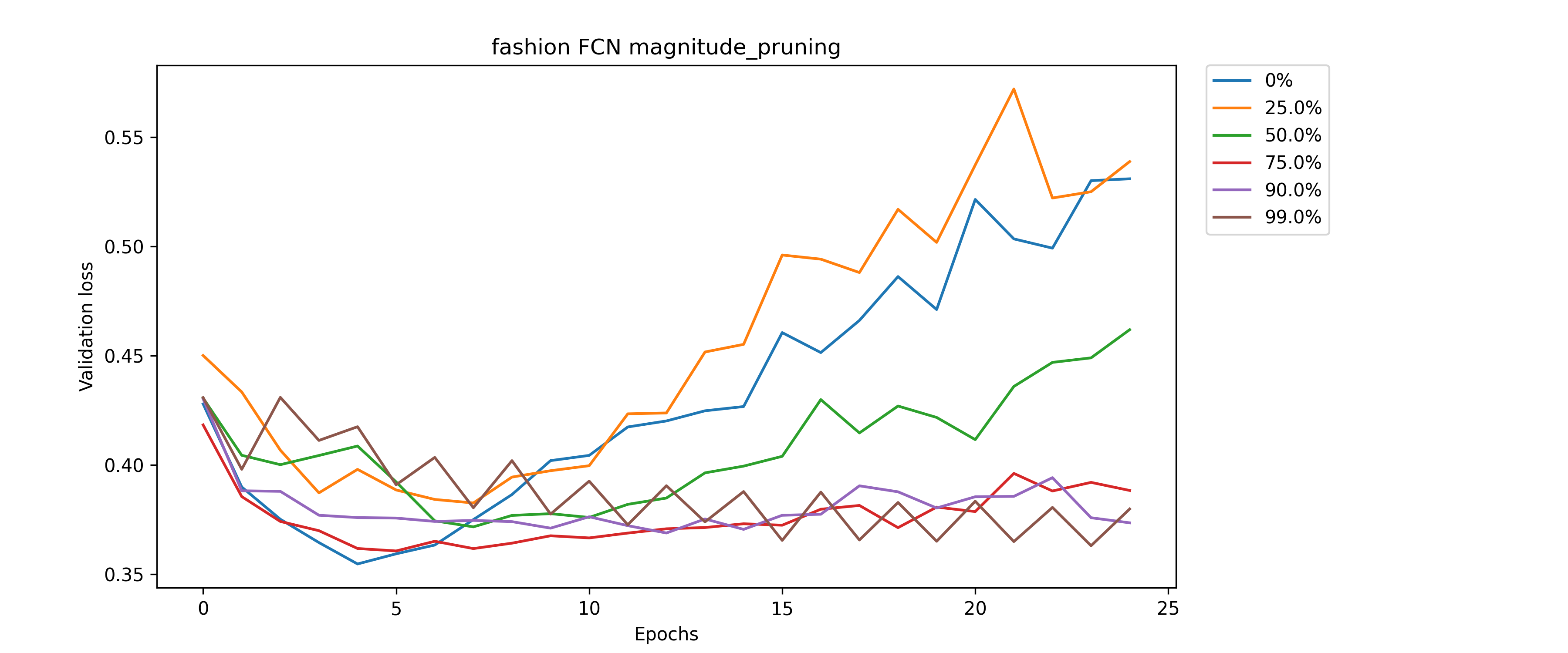}
\caption{Validation loss of magnitude pruning for different sparsity levels.}
\label{fig:fashion_FCN_25_magnitude_pruning_loss_vs_sparsity}
\end{figure}

\FloatBarrier

Figure \ref{fig:fashion_FCN_25_magnitude_pruning_loss_vs_sparsity} shows the validation accuracy of magnitude pruning for different sparsity levels. Higher levels of sparsity improves the validation accuracy of the FCN. The effects of overfitting are reduced as the number of parameters are reduced.

\begin{figure}[h!]
  \centering
  \includegraphics[width=\textwidth]{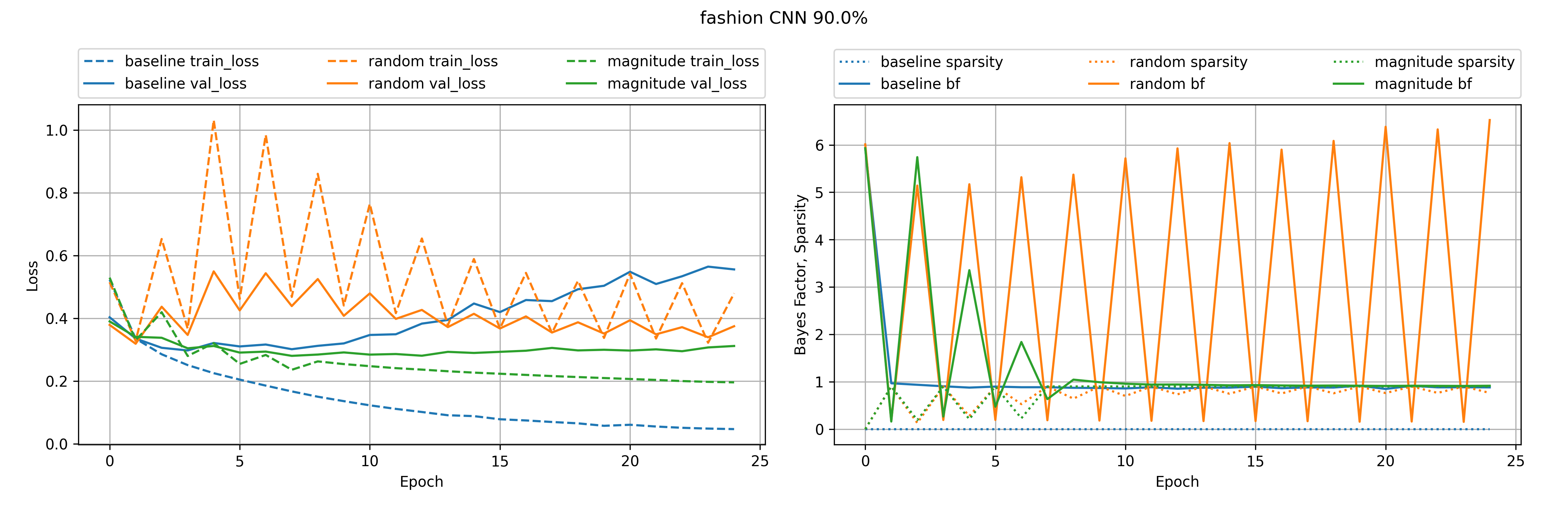}
  \caption{MNIST-Fashion (CNN 90\%) learning curves for the Bayesian pruning method.}
  \label{fig:fashion_CNN_25_0.90_loss}
\end{figure}

\FloatBarrier

Figure \ref{fig:fashion_CNN_25_0.90_loss} shows the learning curves for random pruning, magnitude pruning under a Bayesian framework compared to baseline in a convolutional neural network (CNN) trained on the MNIST Fashion dataset. Here the desired level of sparsity is 90\%. The figure has two subplots. One shows the training and validation loss as a function of the number of epochs, the other plot (right) shows the Bayes factor, sparsity as a function of the number of epochs.

The number of parameters in the CNN are comparatively larger than that of the FCN. This causes the effects of overfitting to be seen a little later in the training period. The trends in the learning curves are similar to that of the FCN. The validation accuracy for random pruning decreases at the beginning of training and starts to improve as training progresses. The Bayes factor begins to reduce as the number of epochs increases and the sparsity of the network becomes stabilized for magnitude pruning, but it remains fluctuating for random pruning and shows an increasing trend for the Bayes factor. Bayesian random pruning model fits the data better than magnitude pruning model.

\begin{figure}[h!]
\centering
\includegraphics[width=\textwidth]{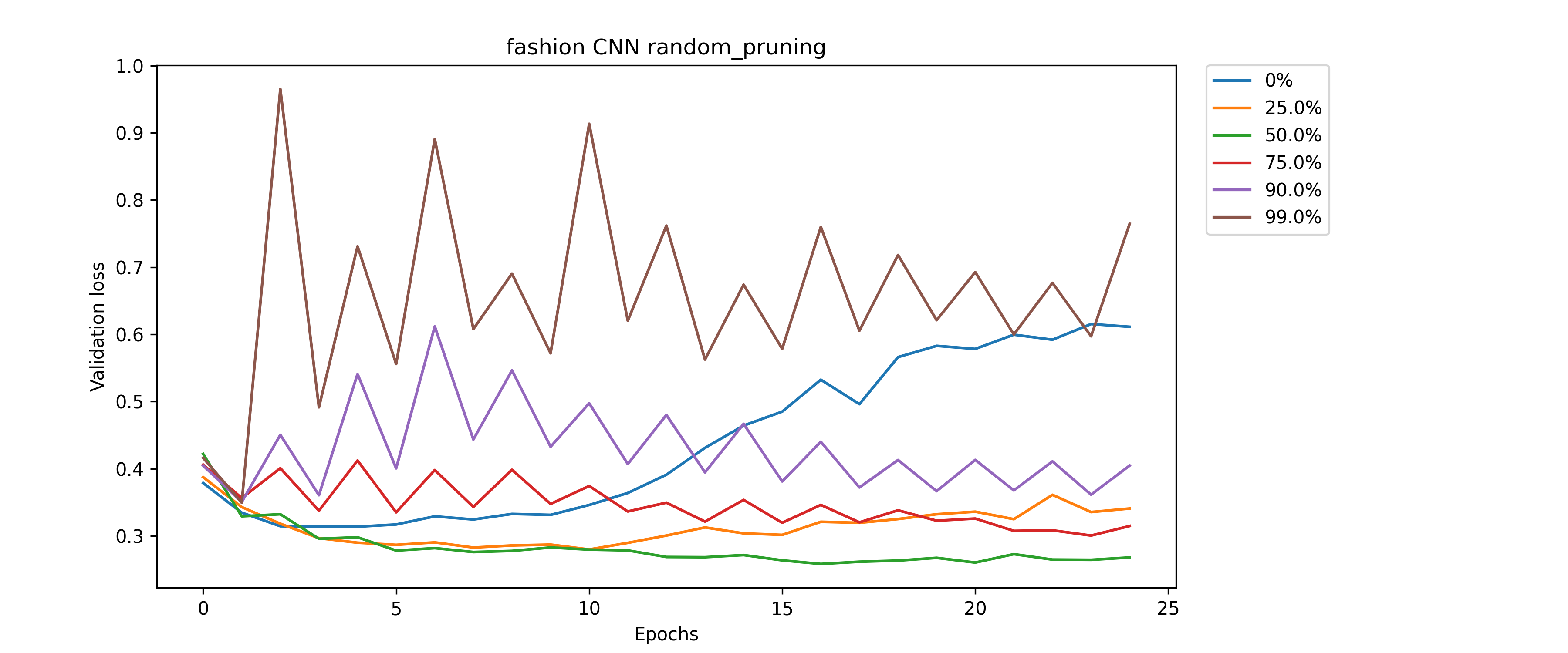}
\caption{Validation loss of random pruning for different sparsity levels.}
\label{fig:fashion_CNN_25_random_pruning_loss_vs_sparsity}
\end{figure}

\FloatBarrier

Figure \ref{fig:fashion_CNN_25_random_pruning_loss_vs_sparsity} shows the validation accuracy of random pruning for different sparsity levels. The trends are similar to the MNIST dataset. The validation accuracy is better for 25\% sparsity and decreases as the sparsity level increases. Sparsity levels up to 90\% helps in reducing the effects of overfitting.

\begin{figure}[h!]
\centering
\includegraphics[width=\textwidth]{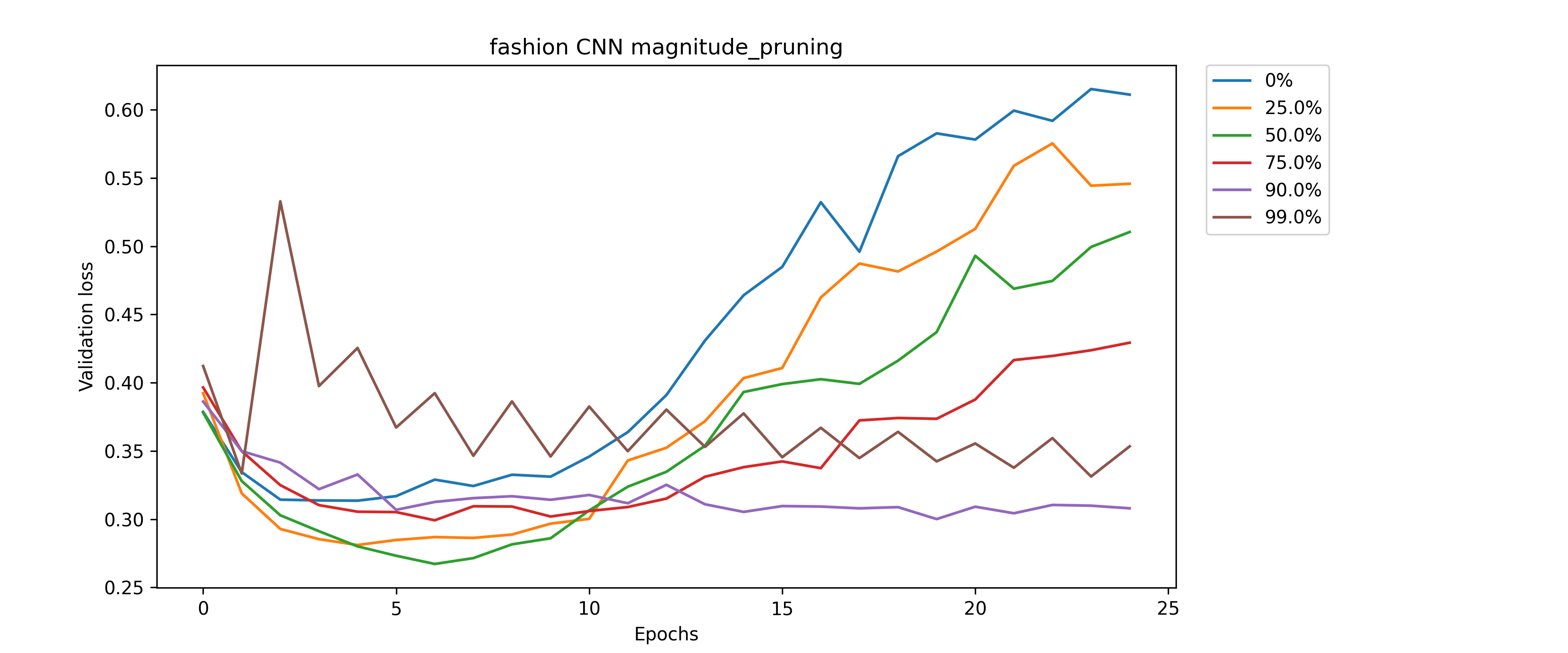}
\caption{Validation loss of magnitude pruning for different sparsity levels.}
\label{fig:fashion_CNN_25_magnitude_pruning_loss_vs_sparsity}
\end{figure}

\FloatBarrier

Figure \ref{fig:fashion_CNN_25_magnitude_pruning_loss_vs_sparsity} shows the validation accuracy of magnitude pruning for different sparsity levels. Similar to the MNIST dataset, magnitude pruning helps in reducing the effects of overfitting. The validation loss continues to improve as 99\% sparsity is achieved.

\subsection*{CIFAR-10}

\begin{figure}[h!]
\centering
\includegraphics[width=\textwidth]{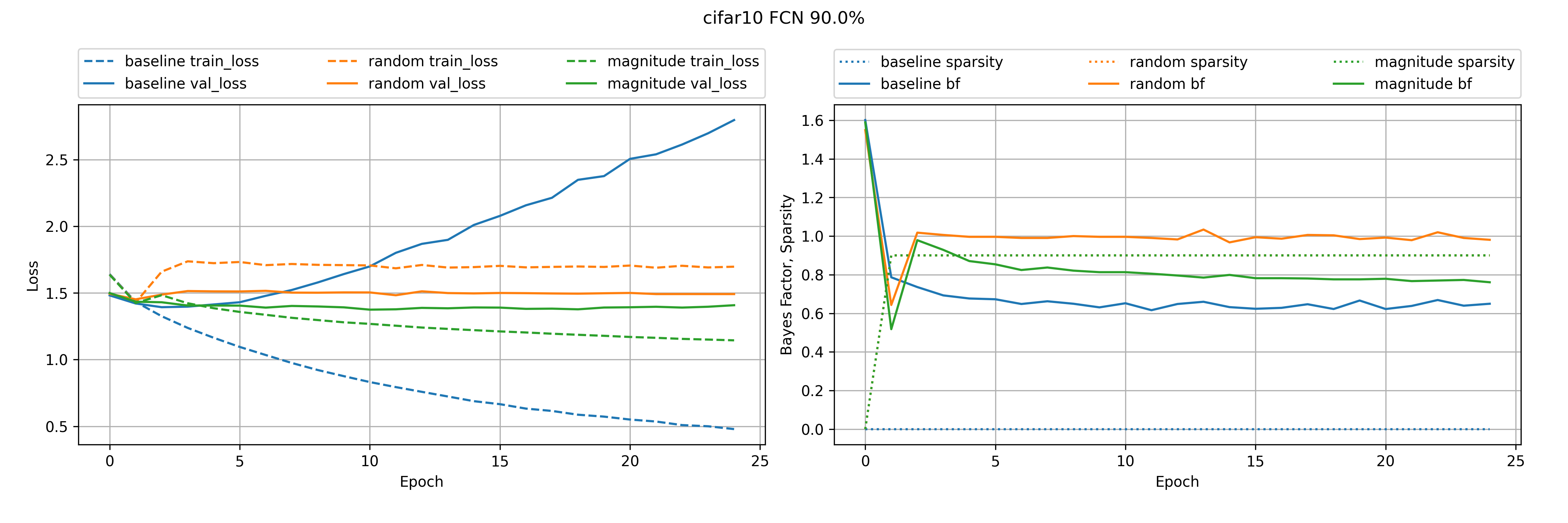}
\caption{CIFAR-10 (FCN 90\%) learning curves for the Bayesian pruning method.}
\label{fig:cifar10_FCN_25_0.90_loss}
\end{figure}

Figure \ref{fig:cifar10_FCN_25_0.90_loss} shows the learning curves for random pruning, magnitude pruning under a Bayesian framework compared to baseline in a fully connected network (FCN) trained on the CIFAR-10 dataset. Here the desired level of sparsity is set to 90\%. The figure has two subplots. One shows the training and validation loss as a function of the number of epochs, the other plot (right) shows the Bayes factor, sparsity as a function of the number of epochs. 

\FloatBarrier

Unlike the MNIST, Fashion datasets the input images of CIFAR-10 dataset are of size 32x32x3. This causes the number of parameters in the FCN to be much larger than that of the MNIST, Fashion datasets. This causes the effects of overfitting to be seen a little later in the training period. The trends in the learning curves are similar to that of the MNIST, Fashion datasets. The validation accuracy for random pruning decreases at the beginning of training and starts to improve as training progresses. The Bayes factor begins to reduce as the number of epochs increases and the sparsity of the network becomes stabilized for both magnitude pruning and random pruning.

\begin{figure}[h!]
\centering
\includegraphics[width=\textwidth]{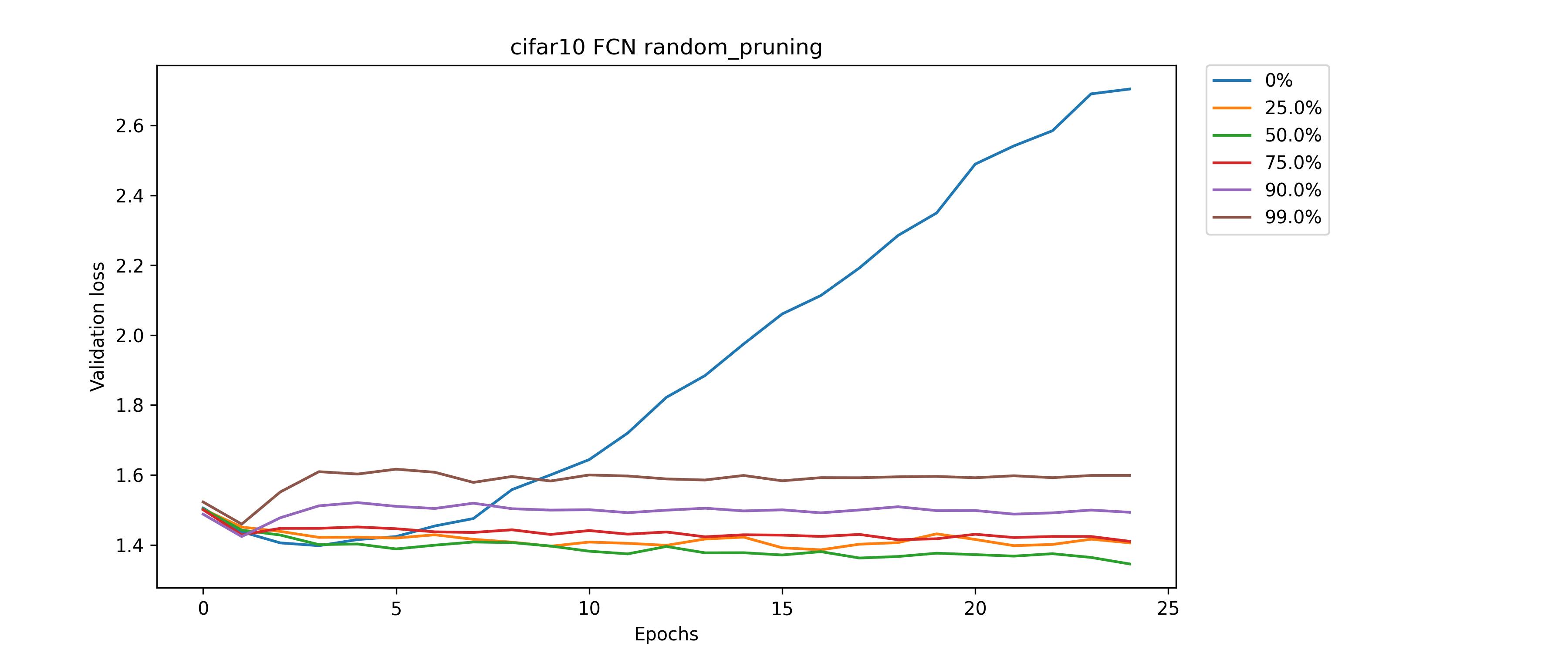}
\caption{Validation loss of random pruning for different sparsity levels.}
\label{fig:cifar10_FCN_25_random_pruning_loss_vs_sparsity}
\end{figure}

\FloatBarrier

Figure \ref{fig:cifar10_FCN_25_random_pruning_loss_vs_sparsity} shows the validation accuracy of random pruning for different sparsity levels. Due to the larger network size, the effects of overfitting are higher. The trends for random pruning remains similar to that of the MNIST, Fashion datasets. The validation accuracy is better for 25\% sparsity and decreases as the sparsity level increases. 

\begin{figure}[h!]
\centering
\includegraphics[width=\textwidth]{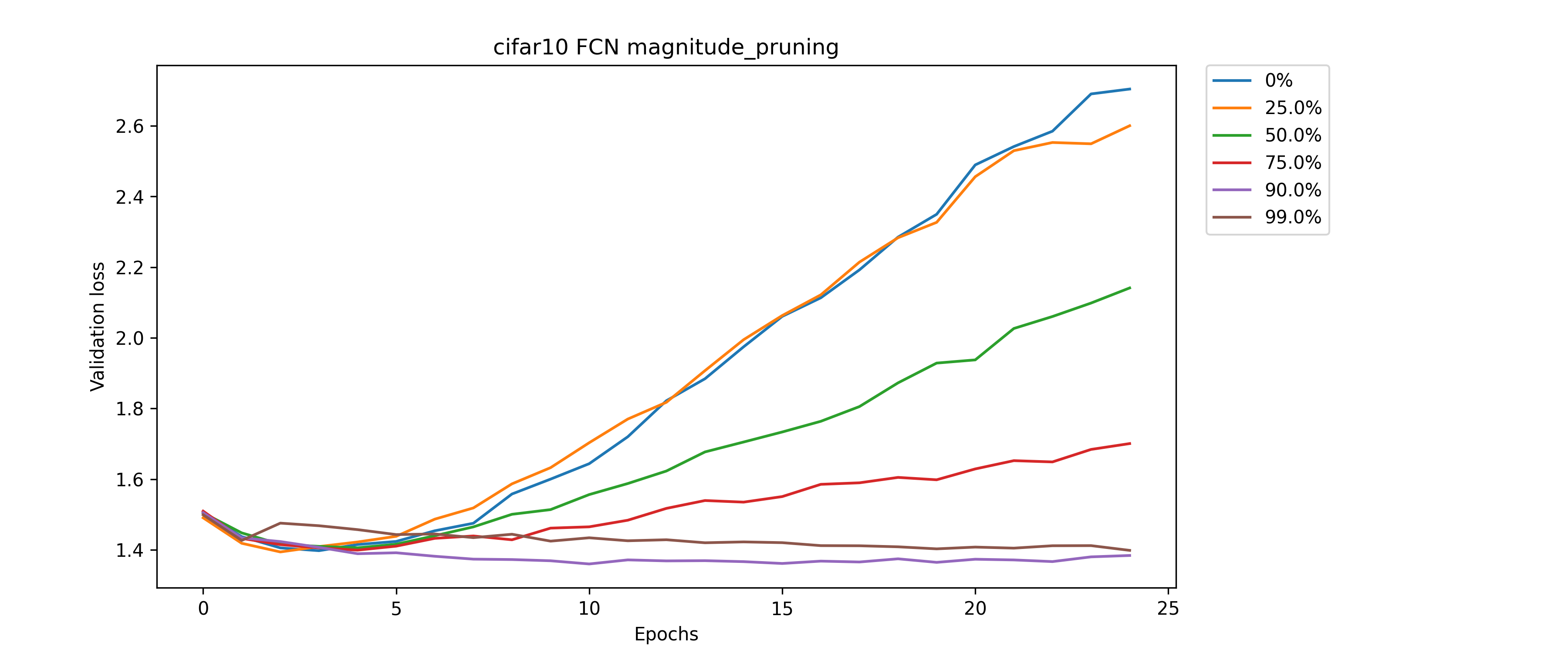}
\caption{Validation loss of magnitude pruning for different sparsity levels.}
\label{fig:cifar10_FCN_25_magnitude_pruning_loss_vs_sparsity}
\end{figure}

\FloatBarrier

Figure \ref{fig:cifar10_FCN_25_magnitude_pruning_loss_vs_sparsity} shows the validation accuracy of magnitude pruning for different sparsity levels. The trends remain the same as that of the MNIST, Fashion datasets. Both Bayesian ranom and Bayesian magnitude pruning helps in reducing the effects of overfitting. The validation loss continues to improve as 99\% sparsity is achieved. Bayesian random pruning model fits the data better than magnitude pruning model.

\begin{figure}[h!]
\centering
\includegraphics[width=\textwidth]{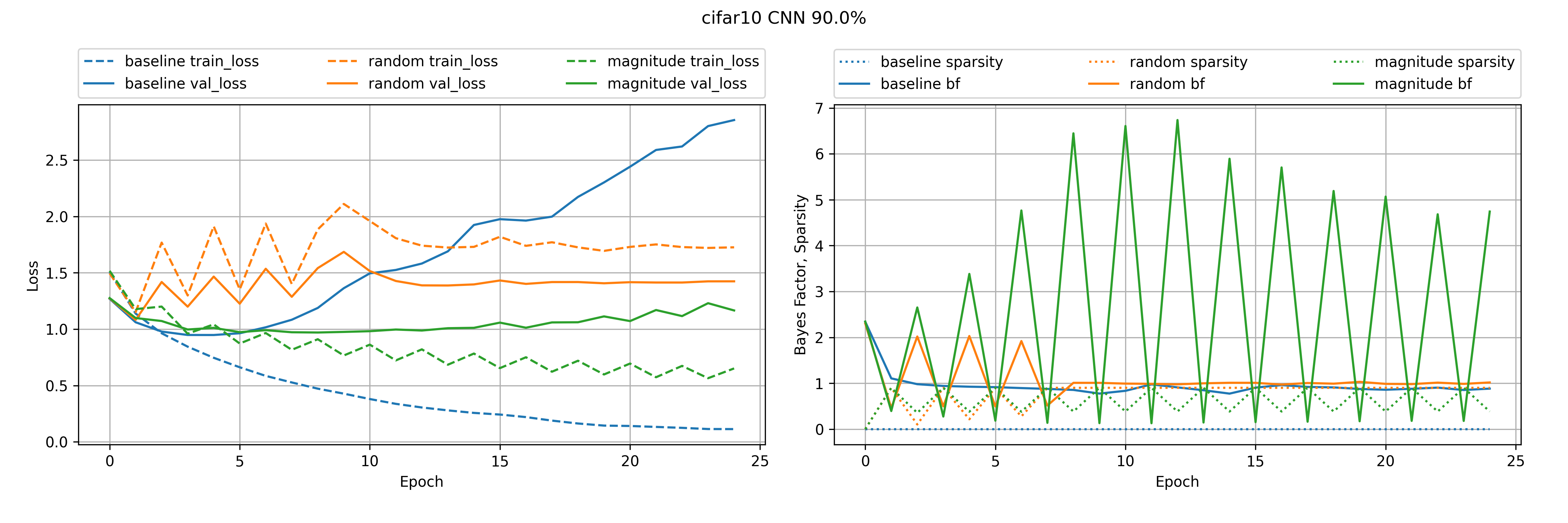}
\caption{CIFAR-10 (CNN 90\%) learning curves for the Bayesian pruning method.}
\label{fig:cifar10_CNN_25_0.90_loss}
\end{figure}

\FloatBarrier

Figure \ref{fig:cifar10_CNN_25_0.90_loss} shows the learning curves for random pruning, magnitude pruning under a Bayesian framework compared to baseline in a convolutional neural network (CNN) trained on the CIFAR-10 dataset. Here the desired level of sparsity is set to 90\%. The figure has two subplots. One shows the training and validation loss as a function of the number of epochs, the other plot (right) shows the Bayes factor, sparsity as a function of the number of epochs. The learning trends are similar to that of the FCN. The validation accuracy for random pruning decreases at the beginning of training and starts to improve as training progresses. The Bayes factor begins to increase for magnitude pruning and sparsity fluctuates as training progresses. For random pruning the Bayes factor begins to reduce as the number of epochs increases and the sparsity of the network becomes stabilized. 

\begin{figure}[h!]
\centering
\includegraphics[width=0.90\textwidth, height=.35\textwidth]{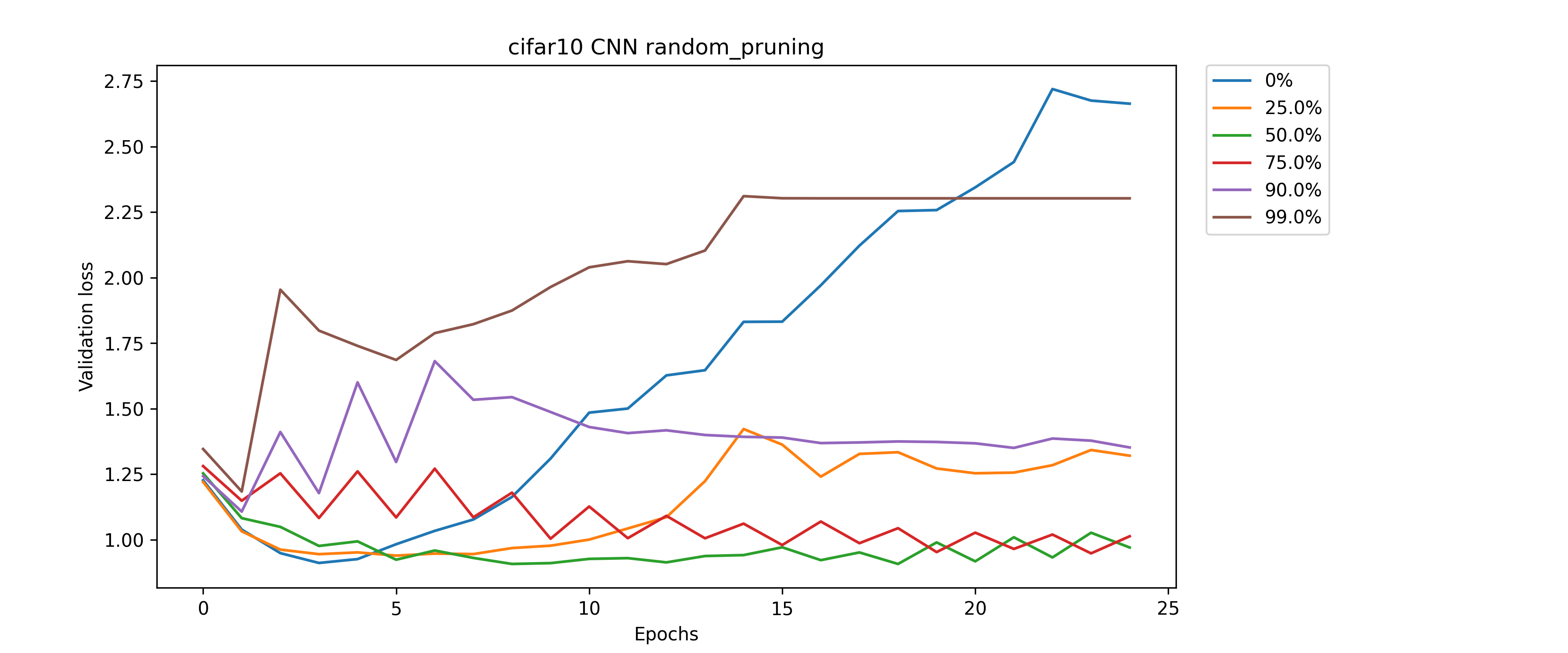}
\caption{Validation loss of random pruning for different sparsity levels.}
\label{fig:cifar10_CNN_25_random_pruning_loss_vs_sparsity}
\end{figure}

\FloatBarrier

Figure \ref{fig:cifar10_CNN_25_random_pruning_loss_vs_sparsity} shows the validation accuracy of random pruning for different sparsity levels. The trends of random pruning is similar to that of the MNIST, Fashion datasets. The effects of overfitting are reduced by pruning. The validation accuracy decreases as the sparsity level increases to 99\%.

\begin{figure}[h!]
\centering
\includegraphics[width=0.90\textwidth, height=.35\textwidth]{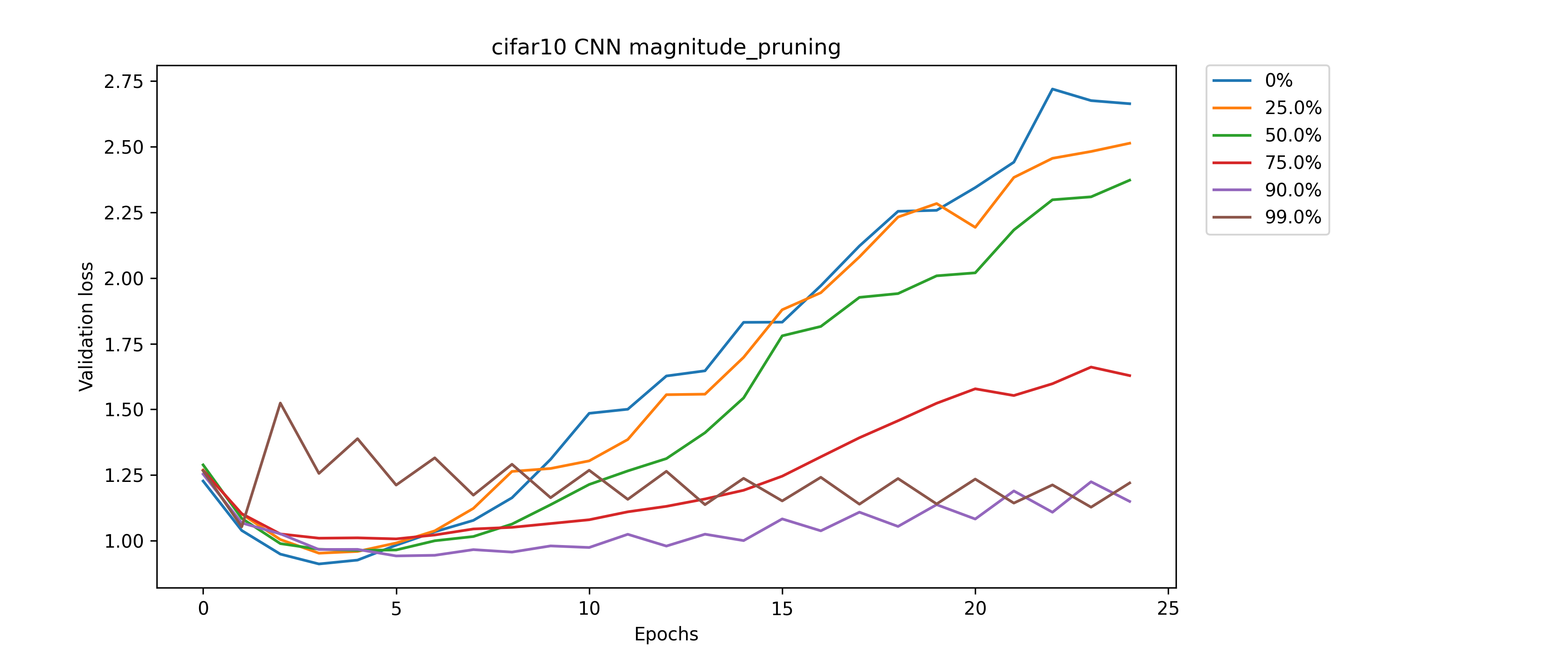}
\caption{Validation loss of magnitude pruning for different sparsity levels.}
\label{fig:cifar10_CNN_25_magnitude_pruning_loss_vs_sparsity}
\end{figure}
\FloatBarrier

Figure \ref{fig:cifar10_CNN_25_magnitude_pruning_loss_vs_sparsity} shows the validation accuracy of magnitude pruning for different sparsity levels. The trends are similar to that of the MNIST, Fashion datasets. Magnitude pruning helps in reducing the effects of overfitting. The validation loss continues to improve as 99\% sparsity is achieved.
\begin{table}[h!]
  \centering
  \caption{Accuracy values at different sparsity levels}
  \adjustbox{max width=\textwidth, max height=0.75\textheight}{%
  \begin{tabular}{cccccccc}
    \toprule
    Dataset & Model & Unpruned & Sparsity & Random & Bayes Random & Magnitude & Bayes Magnitude \\
    \midrule
    \multirow{10}{*}{MNIST} & \multirow{4}{*}{FCN} & \multirow{4}{*}{0.9782} 
    &  25.0\% & 0.9684 & \textbf{0.9747} & \textbf{0.9801} & 0.9759 \\
    &&& 50.0\% & 0.9684 & \textbf{0.9710} & 0.9791 & 0.9791 \\
    &&& 75.0\% & 0.9578 & \textbf{0.9706} & 0.9779 & \textbf{0.9812} \\
    &&& 90.0\% & 0.9624 & \textbf{0.9657} & 0.9768 & \textbf{0.9772} \\
    &&& 99.0\% & 0.9433 & \textbf{0.9439} & 0.9743 & \textbf{0.9767} \\
    \cmidrule{2-8}
    & \multirow{4}{*}{CNN} & \multirow{4}{*}{0.9918} 
    &    25.0\% & \textbf{0.9908} & 0.9835 & 0.9910 & \textbf{0.992} \\
    &&&  50.0\% & 0.9858 & \textbf{0.9906} & 0.9900 & \textbf{0.9901} \\
    &&&  75.0\% & 0.9872 & \textbf{0.9905} & \textbf{0.9905} & 0.9892 \\
    &&&  90.0\% & \textbf{0.9806} & 0.9791 & 0.9880 & \textbf{0.9888} \\
    &&&  99.0\% & 0.1135 & 0.1135 & 0.9826 & 0.9804 \\
    \midrule
    \multirow{10}{*}{Fashion} & \multirow{4}{*}{FCN} & \multirow{4}{*}{0.8733}  
    &   25.0\% & 0.8699 & \textbf{0.8739} & 0.8744 & \textbf{0.8778} \\
    &&&  50.0\% & \textbf{0.8659} & 0.8566 & 0.8725 & \textbf{0.8753} \\
    &&&  75.0\% & 0.8535 & \textbf{0.8558} & \textbf{0.8800} & 0.8799 \\
    &&&  90.0\% & 0.8416 & \textbf{0.8443} & 0.8750 & \textbf{0.8675} \\
    &&&  99.0\% & 0.8076 & \textbf{0.8212} & 0.8573 & 0.8573 \\
    \cmidrule{2-8}
    & \multirow{4}{*}{CNN} & \multirow{4}{*}{0.9028}  
    &   25.0\% & 0.8905 & \textbf{0.9030} & 0.8959 & \textbf{0.9002} \\
    &&&  50.0\% & 0.8957 & \textbf{0.9021} & 0.8906 & \textbf{0.8982} \\
    &&&  75.0\% & 0.8838 & \textbf{0.8773} & 0.8894 & \textbf{0.8974} \\
    &&&  90.0\% & 0.8520 & \textbf{0.8589} & 0.8986 & \textbf{0.9022} \\
    &&&  99.0\% & \textbf{0.7851} & 0.7083 & 0.8595 & \textbf{0.8768} \\
    \midrule
    \multirow{10}{*}{CIFAR-10} & \multirow{4}{*}{FCN} & \multirow{4}{*}{0.4869}  
    &   25.0\% & \textbf{0.5233} & 0.5227 & 0.4857 & \textbf{0.4908} \\
    &&&  50.0\% & \textbf{0.5136} & 0.5111 & 0.4981 & \textbf{0.5010} \\
    &&&  75.0\% & 0.4950 & \textbf{0.4972} & \textbf{0.5109} & 0.5086 \\
    &&&  90.0\% & \textbf{0.4643} & 0.4589 & \textbf{0.5314} & 0.5198 \\
    &&&  99.0\% & 0.4158 & \textbf{0.4381} & \textbf{0.4973} & 0.4932 \\
    \cmidrule{2-8}
    & \multirow{4}{*}{CNN} & \multirow{4}{*}{0.6606} 
    &   25.0\% & 0.6558 & \textbf{0.6574} & 0.6522 & \textbf{0.6557} \\
    &&&  50.0\% & 0.6732 & \textbf{0.6764} & 0.6391 & \textbf{0.6570} \\
    &&&  75.0\% & 0.6205 & \textbf{0.6526} & 0.6409 & \textbf{0.6528} \\
    &&&  90.0\% & \textbf{0.5169} & 0.5092 & \textbf{0.6467} & 0.6437 \\
    &&&  99.0\% & 0.1000 & 0.1000 & 0.5172 & \textbf{0.5537} \\
    \bottomrule
  \end{tabular}%
  }
  \label{tab:bayes_pruning_results}
\end{table}

\FloatBarrier

The accuracy values at different sparsity levels for pruned networks are presented in Table \ref{tab:bayes_pruning_results}. The networks were trained for 25 epochs, and the experiment was repeated 5 times with different random seeds for averaging the results. The table demonstrates that the Bayesian pruning method achieves higher sparsity levels without sacrificing accuracy. It outperforms unpruned networks and shows comparable or better accuracy compared to traditional neural network pruning techniques.

\subsection{Discussion}

Neural networks with a large number of parameters can learn complex functions but are prone to overfitting and are unsuitable for compute-constrained devices. Neural network pruning addresses both these challenges by reducing the network size. The iterative pruning method that we have introduced allows for pruning to a desired level of sparsity without losing any accuracy compared to the baseline. It allows for the network learn a function with fewer connections in principled manner as it checks to see if the network configuration is a good fit for the data. The extensive experiments conducted on three different datasets, two different network types, show that it's an effective method to train neural networks without additional parameterization.

\newpage

\bibliographystyle{Chicago}

\bibliography{bayes_pruning_jcgs}

\end{document}